%% file: tacl2018.tex
\documentclass[11pt,a4paper]{article}
\usepackage[table]{xcolor}
\usepackage{times,latexsym}
\usepackage{url}
\usepackage[T1]{fontenc}

\usepackage{times}
\usepackage{latexsym}
\usepackage{CJKutf8}
\usepackage{xspace,mfirstuc,tabulary}
\usepackage{amsmath,bm}
\usepackage{multirow}
\usepackage{color}
\usepackage{multicol}
\usepackage{tabularx}
\usepackage{booktabs}
\usepackage{pbox}
\usepackage{framed}
\usepackage{array}
\usepackage{enumitem}
\usepackage{chngpage}
\usepackage{threeparttable}
\usepackage{dsfont}
\usepackage{graphicx}
\usepackage{enumitem}
\usepackage{epigraph}
\usepackage{CJKutf8}
\usepackage{amsfonts}
\usepackage{arydshln}
\usepackage{csquotes}
\usepackage{subfig}
\usepackage{circledsteps}

\definecolor{TableShade}{gray}{0.96}
\definecolor{lightgrey}{RGB}{253, 253, 253}
\definecolor{newblue}{RGB}{218, 232, 252}
\definecolor{newgrey}{RGB}{240, 240, 240}

\newcommand{\eg}{{e.g.}}
\newcommand{\ie}{{i.e.}}
\newcommand{\vs}{{vs. }}

\newcommand{\dn}{{C$^3$}} %

\newcommand{\hhide}[1]{}

\usepackage[acceptedWithA]{tacl2018v2}
\usepackage{xspace,mfirstuc,tabulary}

\newif\iftaclinstructions
\taclinstructionsfalse %
\iftaclinstructions
\renewcommand{\confidential}{}
\renewcommand{\anonsubtext}{(No author info supplied here, for consistency with
TACL-submission anonymization requirements)}
\newcommand{\instr}
\fi

\iftaclpubformat %

\else

\fi

\title{Self-Teaching Machines to Read and Comprehend with \\Large-Scale Multi-Subject Question-Answering Data}

\author{
  Dian Yu\textsuperscript{1} ~~Kai Sun\textsuperscript{2}~~ Dong Yu\textsuperscript{1}~~ Claire Cardie\textsuperscript{2} \\
 \textsuperscript{1}Tencent AI Lab, Bellevue, WA \\
 \textsuperscript{2}Cornell University, Ithaca, NY \\
  \{yudian, dyu\}@tencent.com, ks985@cornell.edu, cardie@cs.cornell.edu\\
}

\date{}

\begin{document}
\maketitle

\input{0_abstract}

\input{1_intro}
\input{2_data_generation}

\input{3_method}

\input{4_experiment}

\input{5_related_work}

\input{6_conclusion}

\bibliography{tacl2018}
\bibliographystyle{acl_natbib}

\appendix
\clearpage
\newpage

\input{appendix}

\end{document}

%% file: 0_abstract.tex
\begin{abstract}

In spite of much recent research in the area, it is still unclear whether subject-area question-answering data is useful for machine reading comprehension (MRC) tasks. In this paper, we investigate this question. We collect a large-scale multi-subject multiple-choice question-answering dataset, ExamQA, and use incomplete and noisy snippets returned by a web search engine as the relevant context for each question-answering instance to convert it into a weakly-labeled MRC instance. We then propose a self-teaching paradigm to better use the generated weakly-labeled MRC instances to improve a target MRC task. Experimental results show that we can obtain $+5.1\%$ in accuracy on a multiple-choice MRC dataset, \dn, and $+3.8\%$ in exact match on an extractive MRC dataset, CMRC 2018 over state-of-the-art MRC baselines, demonstrating the effectiveness of our framework and the usefulness of large-scale subject-area question-answering data for different types of machine reading comprehension tasks.

\end{abstract}

%% file: 1_intro.tex
\section{Introduction}

At some level, machine reading comprehension (MRC) and question answering (QA) seem to be quite related tasks: machine reading comprehension aims to answer questions derived from a given document~\cite{richardson2013mctest,hermann2015teaching,rodrigo2015overview}, 
while the standard question answering formulation~\cite{voorhees2000building,burger2001issues,fukumoto2001overview} requires retrieval of snippets of text from a large corpus that answer a given question. 

\begin{figure}[th!]
   \begin{center}
   \includegraphics[width=0.4\textwidth]{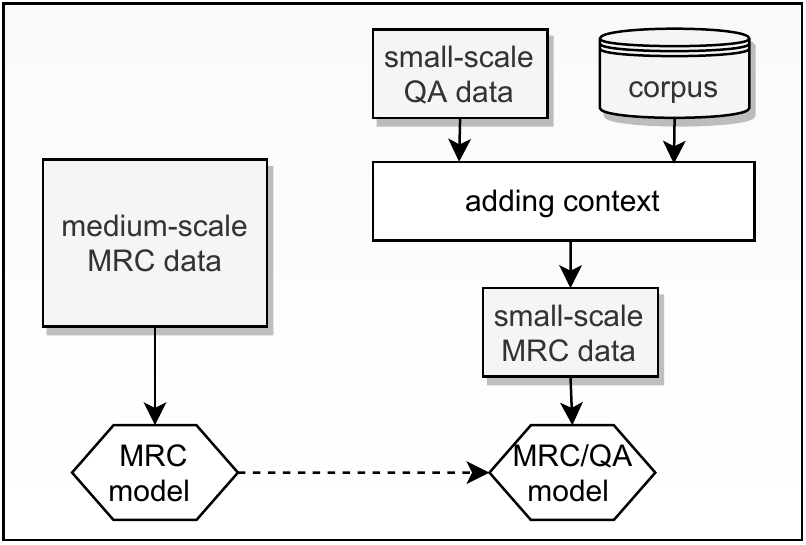}
   \end{center}
 \caption{A typical framework for using medium-scale MRC data to improve small-scale QA.}
 \label{fig:mrc_qa}
\end{figure}

And it has been demonstrated that medium-scale MRC datasets can be employed to improve performance on small-scale question-answering systems.~\newcite{sun2018improving} and \newcite{pan-2019-improving-question}, for example, obtain performance gains on subject-area question-answering datasets about science such as ARC~\cite{clark2016combining,clark2018think} and OpenBookQA~\cite{mihaylov2018can} by pre-training the QA models on MRC data (\eg, RACE~\cite{lai2017race}). The general approach has been to retrieve for each QA instance complete documents or sentences from a relevant in-domain corpus or Wikipedia articles, and then use a machine reading comprehension pipeline on the (context, question) pair to answer questions. See Figure~\ref{fig:mrc_qa} for an overview. Conversely, there actually exists large-scale real-world question-answering data created by subject-matter experts, but rarely is it studied to improve machine reading comprehension.

This paper aims to study whether the massive amount of subject-area QA data can improve machine reading comprehension. For many years it has been demonstrated that human readers' reading comprehension performance is affected by their prior knowledge about the topic of the given text~\cite{johnston1984prior,laufer1985measuring,hirsch2003reading}. Instead of retrieving and imparting topic-specific subject-area knowledge for a given text to a machine reader in an on-demand manner, we hypothesize that incorporating rich knowledge from all --- or as many as possible --- subjects into a machine reader may improve its ability to comprehend text on different topics. As most of the existing multi-subject question-answering datasets are relatively small-scale, we first collect a large-scale \textbf{Q}uestion-\textbf{A}nswering dataset from \textbf{Exam}s covering a wide range of subjects (\eg, sociology, education, and psychology), which contains 638k multiple-choice instances, called \textbf{ExamQA}.

We then present a method to convert QA instances in ExamQA into training instances for a target MRC task, which may benefit from knowledge transfer~\cite{ruder-2019-transfer}. In contrast to previous studies that augment each QA instance with relevant sentences or documents retrieved from offline corpora, we are interested in another practical reading process to add context to QA instances: human readers type their questions on a web search engine and only read through the snippets returned by the search engine to seek potential answers. Imitating this process, we use relevant snippets retrieved by a web search engine as the context of each question-answering instance. We regard such an MRC instance as weakly-labeled as the context is a form of distant supervision: while it might contain the answer to the question as required for MRC, it is equally likely to be noisy, irrelevant, incomplete, and/or too informal to constitute a proper answer (Section~\ref{sec:generation}).

To better leverage the large-scale weakly labeled MRC data, we propose a \textbf{self-teaching} paradigm that iteratively uses a student model that outperforms its teacher model as the new teacher to generate soft labels for data. First, we train a multi-skilled teacher model using both the weakly labeled data and the data of a target MRC task. We then train a multi-skilled student model using the same data while replacing the hard labels of training data with the soft labels predicted by the multi-skilled teacher. Finally, we initialize an expert student model with the resulting multi-skilled student model and fine-tune it on the target MRC data, whose labels are set based on the soft labels generated by the multi-skilled student (Section~\ref{sec:method}). 

We study the effect of the generated QA-based weakly labeled MRC data under the self-teaching paradigm on a multiple-choice MRC dataset, \dn~\cite{sun2019investigating}, in which most questions cannot be solved solely by matching or paraphrasing, and an extractive MRC dataset, CMRC 2018~\cite{cui-2019-span}, in which all answers are spans in the given documents. Experimental results show that we can obtain an $+5.1\%$ in accuracy on \dn~and $+3.8\%$ in exact match on CMRC 2018 over state-of-the-art baselines~\cite{xu2020clue,cui-2020-revisiting}. Furthermore, we present an easy way to adapt this paradigm to additionally leverage multiple types of weakly labeled MRC data wherein noise is introduced by different factors (\eg, context retrieval, machine translation, and knowledge construction), again by using soft labels predicted by teacher models. Augmenting the MRC training data in this way leads to further improvements (up to $+2.5\%$ in accuracy on \dn). These results demonstrate the effectiveness and flexibility of our paradigm.

The contributions of this paper are as follows.
\begin{itemize}
\setlength\itemsep{-0.1em}
    \item We offer the largest multi-subject QA dataset to date, ExamQA. ExamQA will be available at \url{https://dataset.org/examqa/}. 
    \item Our study examines whether large-scale subject-area QA data can improve MRC. We explore an approach that adds noisy and incomplete context retrieved by a web search engine to QA instances to convert them into weakly-labeled MRC data.
    \item We propose a simple yet effective self-teaching paradigm to better leverage large-scale weakly-labeled data of one or multiple types (\ie, in which the source of noise varies). Experimental results show that we can achieve up to $+7.6\%$ in accuracy on a challenging multiple-choice MRC dataset and $+3.8\%$ in exact match on a representative extractive MRC dataset. 
\end{itemize}

%% file: 2_data_generation.tex
\section{Weakly-Labeled Data Generation}
\label{sec:generation}

\subsection{Question-Answering Data Collection}

We collect large-scale question-answering instances from freely accessible exams (including mock exams) designed for a variety of subjects such as programming, journalism, and ecology. We only keep multiple-choice single-answer instances written in \textbf{Chinese}. After deduplication, we obtain 638,436 question-answering instances. 

To assess the subject coverage of ExamQA, we obtain a list of subjects from China national standard (GB/T 13745-2009)~\cite{GBT137452009} and check for each subject in the list if the name of the subject appears in the title of any exam to estimate the lower bound of subject coverage. The estimation shows that ExamQA covers \textbf{at least} 48 out of 62 first-level subjects and 187 out of 676 second-level subjects. Note that the actual subject coverage of ExamQA may be greatly underestimated, as only $24.2\%$ of titles contain a subject name.

We do not annotate a small subset of questions for human performance, as most of the subject-area questions are from higher education exams that require advanced domain knowledge.

\subsection{Comparisons with Existing Subject-Area Question-Answering Datasets}

Subject-area question answering is an increasingly popular direction in question answering, focusing on closing the performance gap between humans and machines in answering questions collected from real-world exams that are carefully designed by subject-matter experts. These tasks are mostly in multiple-choice forms. In Table~\ref{tab:datasets}, we list several representative subject-area multiple-choice question-answering datasets: NTCIR-11 QA-Lab~\cite{shibuki2014overview}, QS~\cite{chengtaking2016}, MCQA~\cite{guo-etal-2017-ijcnlp}, ARC~\cite{clark2018think}, GeoSQA~\cite{huang2019geosqa}, HEAD-QA~\cite{vilares-gomez-rodriguez-2019-head}, EXAMS~\cite{hardalov-etal-2020-exams}, JEC-QA~\cite{zhong2020jec}, and MEDQA~\cite{jin2020disease}.

\begin{table}[h!]
\centering
\scriptsize
\begin{tabular}{llllll}
\toprule
\bf dataset   & \bf \# of subjects$^\circ$ & \bf subjects  & \bf size \\   %
\midrule
QS                  & 1 &history               & 0.6K    \\   %
GeoSQA                 & 1 &geography                 & 4.1K     \\ %
JEC-QA                  & 1 &legal                & 26.4K    \\  %
ARC                        &  1 &science              & 7.8K   \\   %
QA-Lab                  & 1 &history    & 0.3K     \\    %
HEAD-QA                & 1 &healthcare         & 6.8K   \\  %
MEDQA                     & 1 &medical          & 61.1K  \\ %
MCQA        & 6 &multi-subject                 & 14.4K   \\ %
EXAMS       & 24 &multi-subject                & 24.1K    \\ %
\midrule
\bf ExamQA     & 48 &multi-subject             & 638.4K  \\ %
\bottomrule
\end{tabular}
\caption{\label{tab:datasets} Representative subject-area QA datasets collected from exams ($^\circ$: we simply report the number of subjects stated by previous studies and the number of first-level subjects in ExamQA).}
\end{table}

It is worth mentioning that some multiple-choice MRC datasets (\eg,~\cite{rodrigo2015overview,lai2017race}) are collected from language exams designed to test the reading comprehension ability of a human reader. These kinds of context-dependent problems are \textbf{not} included in ExamQA.

\subsection{Bringing Context to Question Answering}

In this section, we present a method to convert QA instances into multiple-choice or extractive MRC instances to make the resulting data and target MRC task in a similar format, which may benefit from knowledge transfer~\cite{ruder-2019-transfer}. 

Previous studies attempt to convert a multiple-choice subject-area QA task to a multiple-choice MRC task by retrieving relevant sentences for each question from a clean corpus to form a document. 
Instead of relying on a relatively clean resource, we retrieve the top ranked snippets using a publicly available search engine. Specifically, we send each question to the search engine as the query and collect snippets from the first result page. Typically, we can collect ten snippets for each QA instance. Since all instances are freely accessible online, it is likely that a retrieved snippet merely contains the original QA instance rather than relevant context sufficient for answering the question. Therefore, we discard a snippet if more than one answer option appears as a substring in the snippet. We concatenate the remaining snippets into a document as the context of each QA instance. We show data statistics of ExamQA and retrieved context in Table~\ref{tab:examqastat}. Due to this construction method, it is very likely that a document is noisy, incomplete, informal, and/or irrelevant. We provide sample instances in Table~\ref{tab:sample1b-en}.

To convert these multiple-choice MRC instances into extractive ones, we remove the wrong answer options of each multiple-choice instance and append the start/end offsets of the first mention of the correct answer in its associated snippet.

\begin{table}[h!]
\centering
\scriptsize
\begin{tabular}{ll}
\toprule
\bf metric                & \bf value  \\
\midrule
average \# of answer options &  4.0      \\ %
average question length (in characters) & 39.5 \\
average option length (in characters) & 6.7 \\
average context length (in characters) & 907.6 \\
character vocabulary size & 13,258 \\
non-extractive correct option (\%) & 68.4 \\
\bottomrule
\end{tabular}
\caption{\label{tab:examqastat} Data statistics of ExamQA with context.}
\end{table}

\begin{table}[h!]
\centering
\scriptsize
\begin{tabular}{p{0.2cm}p{6.5cm}}
\toprule
\textbf{C1}: & 1. + b / b is equivalent to ((int) a) + (b / b), which can be obtained according to the priority of the processor. (Int) This is a forced type conversion. After the forced conversion ((int) a) is generally the double conversion to the int type, most platforms round to zero... 2./b, both sides of the division sign are doubletype , The result is also doubleType. That is 1.000000; integer. The first 5 is the int type, int... 3 .; a = 5.5; b = 2.5; c = (int) a + b / b; printf (\" .. Best answer: (int) a + b / b = 6, should be (int) a means round a, and round a is 5 (rounding cannot be used here, rounding is discarded, then b / b is 2.5 / 2.5, etc... 2019 July 25th, 2016-Analysis: The type of the value of the mixed expression is determined by the type with the highest precision in the expression, so it can be seen that option B can be excluded. Note that the result of b / b should be 1.00000, and (int) a is 5, and the result of the addition is still double...\\

\textbf{Q1}: & Suppose a and b are double constants, a=5.5, and b=2.5, the value of the expression (int)a+b/b is ().  \\
\multicolumn{2}{l}{\hspace{7mm}A.\hspace{2mm}5.500000.} \\
\multicolumn{2}{l}{\hspace{7mm}B.\hspace{2mm}6.000000. $\star$} \\
\multicolumn{2}{l}{\hspace{7mm}C.\hspace{2mm}6.500000.}\\
\multicolumn{2}{l}{\hspace{7mm}D.\hspace{2mm}6.}\\

\midrule
\textbf{C2}: & November 22, 2016 It can be seen that it is not a white box test case design method, so the correct answer to question (31) is B. Black box testing is also called functional testing, which is to detect whether each function can be used normally. At the test site, treat the program as... November 18, 2016 Black box testing technology is also called functional testing, which tests the external characteristics of the software without considering the internal structure and characteristics of the software. The main purpose of black box testing is to discover the following types of errors: Are there any errors... [Answer Analysis]... \\
\textbf{Q2}: &  Black box testing is also called functional testing, and black box testing cannot find ().\\
\multicolumn{2}{l}{\hspace{7mm}A.\hspace{2mm}terminal error.} \\
\multicolumn{2}{l}{\hspace{7mm}B.\hspace{2mm}communication error.}\\
\multicolumn{2}{l}{\hspace{7mm}C.\hspace{2mm}interface error.}\\
\multicolumn{2}{l}{\hspace{7mm}D.\hspace{2mm}code redundancy. $\star$}\\

\midrule
\textbf{C3}: & July 21, 2014-Friedman believes that the transmission variable of monetary policy should be (). Please help to give the correct answer and analysis, thank you! Reward: 0 answer bean Questioner: 00***42 Release time: 2014-07-21 View... \\
\textbf{Q3}: & Friedman believes that the transmission variable of monetary policy should be ().\\
\multicolumn{2}{l}{\hspace{7mm}A.\hspace{2mm}excess reserve.} \\
\multicolumn{2}{l}{\hspace{7mm}B.\hspace{2mm}interest rate.}\\
\multicolumn{2}{l}{\hspace{7mm}C.\hspace{2mm}currency supply. $\star$}\\
\multicolumn{2}{l}{\hspace{7mm}D.\hspace{2mm}base currency.}\\

\bottomrule
\end{tabular}

\caption{English translation of sample instances in ExamQA with retrieved context ($\star$: the correct answer option).
}
\label{tab:sample1b-en}
\end{table}

%% file: 3_method.tex
\section{Self-Teaching Paradigm}
\label{sec:method}

In this section, we introduce a self-teaching paradigm to leverage large-scale QA-based weakly-labeled data to improve the performance of existing supervised methods on an MRC task of interest, which is relatively small-scale. Due to limited space, here we only discuss multiple-choice data and tasks and leave the reformulation (\eg, soft labels and loss functions) for extractive MRC tasks in Appendix~\ref{sec:appendix}.

\subsection{Training a Multi-Skilled Teacher}
\label{sec:method:multi-teacher}

In previous teacher-student frameworks~\cite{you2019teach,wang2019go,sun2020improving}, multiple teacher models are trained using different data. However, it is difficult to divide the weakly-labeled data based on existing QA instances into sub-datasets by subjects or fine-grained types of knowledge required for answering questions. Instead, we train a multi-skilled teacher model using both the human-annotated target MRC data and the weakly-labeled data, which requires a diverse skill set and knowledge from a variety of domains.

Let $V$ denote a set of human-annotated instances and $W$ denote a set of weakly-labeled instances. For each instance $t \in V\cup W$, we let $m_t$ denote its total number of answer options, and ${\bm h}^{(t)}$ be a one-hot (hard-label) vector such that ${h}^{(t)}_j = 1$ if the $j$-th answer option is labeled as correct. We train a single multi-skilled teacher model, denoted by $\mathcal{T}$, and optimize $\mathcal{T}$ by minimizing $\sum_{t \in V\cup W} L_1(t, \theta_{\mathcal{T}})$; $L_1$ is defined as 

{\small
\begin{equation*}
L_1(t, \theta) = - \sum_{1 \le k \le m_t} {h}^{(t)}_k ~ \log p_\theta(k\,|\,t), 
\end{equation*}
}%
where $p_\theta(k\,|\,t)$ denotes the probability that the $k$-th answer option of instance $t$ is correct, estimated by the  model with parameters $\theta$.

\begin{figure*}[ht!]
   \begin{center}
   \includegraphics[width=0.96\textwidth]{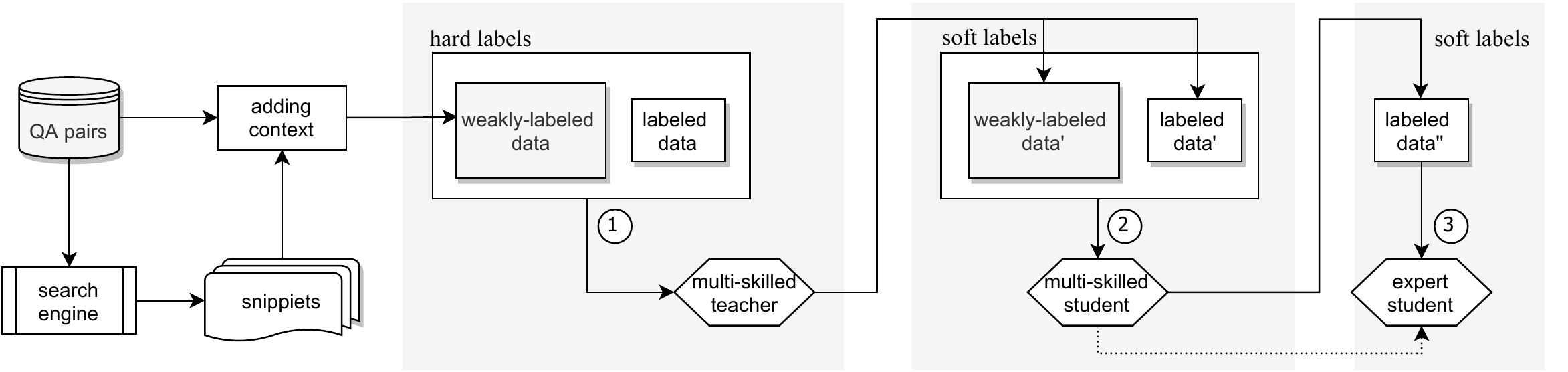}
   \end{center}
 \caption{Self-teaching framework using large-scale QA data to improve relatively small-scale MRC.}
 \label{fig:overview_single}
\end{figure*}

\subsection{Training a Multi-Skilled Student}
\label{sec:method:multi-student}

We then train a multi-skilled student model $\mathcal{S}$ using the same data as the multi-skilled teacher model $\mathcal{T}$ while replacing the hard labels of answer options with the soft labels predicted by $\mathcal{T}$. We define soft-label vector ${\bm s}^{(t)}$ for $t \in V\cup W$ such that
{\small
\begin{equation*}
    {s}^{(t)}_k =
     \lambda~{h}^{(t)}_k + (1 - \lambda) p_{\theta_{\mathcal T}}(k\,|\,t)  \\,
\end{equation*}
}%
where $\lambda\in [0, 1]$ is a weight parameter, and $k = 1,\dots, m_t$. 

We optimize multi-skilled student $\mathcal{S}$ by minimizing $\sum_{t \in V\cup W}L_2(t, \theta_{\mathcal S})$, where $L_2$ is defined as
{\small
\begin{equation*}
    L_2(t, \theta) = -\sum_{1\le k\le m_t} {s}^{(t)}_k ~ \log p_{\theta}(k\,|\,t).
\end{equation*}
}%

\subsection{Training an Expert Student}
\label{sec:method:expert-student}

Finally, we initialize an expert student $\mathcal{E}$ with the resulting multi-skilled student model $\mathcal{S}$, and we fine-tune $\mathcal{E}$ on the target data $V$ to help it achieve expertise in the task of interest, following most of the recent MRC methods~\cite{radfordimproving,devlin-etal-2019-bert}. This step differs from previous work in that we use the soft labels generated by the multi-skilled student model (Section~\ref{sec:method:multi-student}) based on our assumption that a student model tends to learn better from a stronger teacher model. We will discuss more details in the experiment section and show that a student model tends to outperform its teacher model that provides soft labels to make itself a stronger teacher (Section~\ref{sec:expriment}).

We define new soft-label vector ${\bm{\tilde{s}}}^{(t)}$ for $t \in V$ such that
{\small
\begin{equation*}
    {\tilde s}^{(t)}_k =
     \lambda~{h}^{(t)}_k + (1 - \lambda) p_{\theta_{\mathcal S}}(k\,|\,t)  \\,
\end{equation*}
}%
where $\lambda\in [0, 1]$ is a weight parameter, and $k = 1,\dots, m_t$. 

In this stage, we optimize $\mathcal{E}$ by minimizing $\sum_{t\in V} L_3(t, \theta_{\mathcal E})$, where $L_3$ is defined as

{\small
\begin{equation*}
    L_3(t, \theta) = -\sum_{1\le k\le m_t} {\tilde{s}}^{(t)}_k ~ \log p_{\theta}(k\,|\,t).
\end{equation*}
}%

Figure~\ref{fig:overview_single} shows an overview of the proposed self-teaching paradigm.

\subsection{Integrating Different Types of Weakly-Labeled Data}
\label{sec:method:integration}

We study the integration of multiple types of weakly-labeled data during weakly-supervised training with soft labels to save time and effort in retraining models on $W$ with hard labels.

Take another weakly-labeled multiple-choice MRC data extracted automatically from television show and film \textbf{s}cripts~\cite{sun2020improving} as an example, denoted as $W_s$, besides the weakly-labeled data $W$ we construct based on existing question-answering instances. Following the above three-step procedure, we first train a multi-skilled teacher $\mathcal{T}_s$ using $W_s$ to generate soft labels of $W_s$ and $V$. We then train a multi-skilled student $\mathcal{S}_{\ast}$ upon the combination of soft-labeled $W_s$, $W$ (Section~\ref{sec:method:multi-student}), and $V$. Note that we simply use two versions of soft-labeled $V$ generated by $\mathcal{T}$ and $\mathcal{T}_s$, respectively. The resulting student $\mathcal{S}_{\ast}$ is used to generate the final soft labels of $V$ for training an expert student. We will discuss integration with other types of weakly-labeled data wherein noise is introduced by various factors in Section~\ref{sec:exp:main}.

%% file: 4_experiment.tex
\section{Experiments}
\label{sec:expriment}

\subsection{Data Statistics}

We show statistics of two relatively small-scale target MRC datasets and two kinds of large-scale weakly-labeled MRC data in Table~\ref{tab:exp:statistics}. Note that for CMRC 2018, we use its publicly available training and development sets.

\begin{table}[ht!]
\centering
\scriptsize
\begin{tabular}{p{1.8cm}ll}
\toprule
\bf data          & \bf source            & \bf \# of instances \\
\midrule
\multicolumn{3}{l}{\textbf{human-annotated}:} \\ 
\dn     & language exams & 19,577  \\
CMRC 2018   & Wikipedia                  & 19,071 \\
\midrule
\multicolumn{3}{l}{\textbf{weakly-labeled}:} \\ 
SCRIPT      & TV/movie scripts & 700,816  \\
ExamQA    & multi-subject exams  & 638,436  \\
\bottomrule
\end{tabular}
\caption{\label{tab:exp:statistics} Human-annotated and weakly-labeled machine reading comprehension data statistics.}
\end{table}

\begin{table*}[ht!]
\centering
\scriptsize
\begin{tabular}{llllllcc}
\toprule
\multirow{2}{*}{\bf id} & \multirow{2}{*}{\bf model} & \multirow{2}{*}{\bf init.} & \multirow{2}{*}{\bf teacher}  & \multicolumn{2}{c}{\bf training data} & \multirow{2}{*}{\bf dev} & \multirow{2}{*}{\bf test} \\
\cline{5-6}
    &   &  &   & \bf name       & \bf label       &      &                       \\
\midrule
0       & RoBERTa-wwm-ext-large~\cite{xu2020clue}     &    --    & -- &  $\diamondsuit$    & hard  & --    & 73.8 \\
\midrule
1     & baseline (our implementation of 0)           &    --     & -- &  $\diamondsuit$    & hard  & 73.9  \scriptsize(0.5)    & 73.4 \scriptsize(0.5)                       \\
2     & multi-skilled teacher  & -- & -- & $\diamondsuit$ + ExamQA  & hard  & 74.0 \scriptsize(0.8)    & 75.6 \scriptsize(0.5)     \\   
3     & multi-skilled student  & -- & 2 &  $\diamondsuit$ + ExamQA  & soft   & 75.7 \scriptsize(0.5)    & 77.1 \scriptsize(0.4)    \\  
4     & expert student (variant)       & 3  & 2 &  $\diamondsuit$ & soft   & 77.8 \scriptsize(0.4) &  78.0 \scriptsize(0.3)\\
5     & expert student       & 3   & 3 &  $\diamondsuit$ & soft   & \bf 78.2 \scriptsize(0.3)    & \bf 78.5 \scriptsize(0.2)\\
\bottomrule
\end{tabular}
\caption{\label{tab:performance} Average accuracy and standard deviation ($\%$) on the dev and test sets of the {\dn} dataset. $\diamondsuit$ is the training set of \dn~for all experiments; init. means the starting point, and -- in this column means using the pre-trained language model for initialization.}
\end{table*}

\subsection{Implementation Details}
We follow recent state-of-the-art MRC methods for the model architecture that consists of a pre-trained language model and a classification layer. We use the same architecture for baselines and all teacher or student models. We use RoBERTa-wwm-ext-large~\cite{cui-2020-revisiting} as the pre-trained language model for Chinese, which achieves state-of-the-art performance on representative MRC tasks such as {\dn} and CMRC 2018~\cite{xu2020clue}. We are aware of the emerging new pre-trained language models for Chinese and leave the exploration of them for future studies. We train a multi-skilled teacher or student model for one epoch as large-scale weakly-labeled data is used. We train baselines and expert students on {\dn} and CMRC 2018 for eight epochs and two epochs, respectively. In all experiments, we set $\lambda$ (defined in Section~\ref{sec:method:multi-student}-\ref{sec:method:expert-student}) to $0.5$  to allow easy comparisons with previous work~\cite{sun2020improving}, and we report the average score of five runs with different random seeds and standard deviation in brackets. See more setting details in Appendix~\ref{sec:appendix:settings}.

\subsection{Main Results and Observations}
\label{sec:exp:main}

As shown in Table~\ref{tab:performance}, under the proposed self-teaching paradigm, expert student (5) improves the state-of-the-art baseline (1) based on the same model architecture by up to $5.1\%$ in accuracy on the \dn~dataset. We also compare (5), its variant (4), and the intermediate teacher models (2 and 3), and we have the following observations. 

$\\$
\noindent \textbf{Student models tend to outperform their corresponding teacher models in self-teaching.}
Under the self-teaching paradigm, we notice that student models always achieve higher accuracy on the MRC task of interest than their teacher models that generate soft labels for them. For example, the accuracy of the multi-skilled student (3) is $1.5\%$ higher than the result $75.6\%$ achieved by the multi-skilled teacher (2).

$\\$
\noindent \textbf{Using a strong multi-skilled model to provide soft labels helps across settings.} We consider a teacher model to be strong if it achieves good performance on the MRC task of interest. We already demonstrate that training a student model with soft labels generated by a multi-skilled teacher model instead of hard labels yields positive improvements (3 \vs2). In addition, using the multi-skilled student (3), which is stronger than the multi-skilled teacher (2), to provide soft labels of \dn~to train an expert student results in a $0.5\%$ increase in accuracy (5 \vs4).

To explore whether this also applies to expert models, we experiment with a variant of expert student (5): still starting from the same multi-skilled student (3), we now put back expert student (5) as the teacher model to generate soft labels of \dn~to train an expert student variant. However, this variant does not yield further gains ($78.2$ ($\scriptsize 0.4$)) on the development set). Seeing more data than the expert student may make the more \emph{``knowledgable''} multi-skilled student a better teacher to provide soft labels of the target MRC data. While it is possible to use the multi-skilled student itself to obtain a stronger multi-skilled student, it is much less efficient to retrain a model upon the large-scale weakly-labeled data than the above variant. We leave the exploration of iterative self-teaching over weakly-labeled data to future work.

$\\$
\noindent \textbf{Large-scale weakly-labeled data based on subject-area QA instances can be helpful for MRC.}
We do not see noticeable improvements by merely combining large-scale weakly-labeled data and small-scale data to train a multi-skilled teacher over the resulting data for one epoch (2 in Table~\ref{tab:performance}). Nevertheless, helping train multi-skilled teacher/student models, especially the multi-skilled student that is further used as a good starting point of the expert student, reflect the usefulness of the large-scale weakly-labeled data. Though starting from the multi-skilled teacher slightly boosts ($+0.3\%$ in accuracy) a multi-skilled student's performance (Table~\ref{tab:ablation}), using the resulting multi-skilled student to initialize and teach the expert student actually hurts performance ($-0.7\%$ in accuracy on the development set), perhaps due to the overuse of the same weakly-labeled data (both soft and hard labels) upon a single model or the convergence between the multi-skilled teacher and multi-skilled student. Therefore, we do not use the multi-skilled teacher to initialize the multi-skilled student in our main experiment (3 in Table~\ref{tab:performance}).

\begin{table}[h!]
\centering
\scriptsize
\begin{tabular}{llc}
\toprule
\bf model & \bf initialization & \bf dev \\
\midrule
expert student             & multi-skilled student        & \textbf{78.2} \scriptsize(0.3)     \\
expert student             & multi-skilled teacher         & 77.7 \scriptsize(0.3)       \\  
expert student             & --         & 74.9 \scriptsize(0.3)       \\   
\midrule
multi-skilled student      & multi-skilled teacher       & 76.0 \scriptsize(0.2)            \\
multi-skilled student      & --         & 75.7 \scriptsize(0.5)       \\

\bottomrule
\end{tabular}
\caption{\label{tab:ablation} Ablation study and variant comparisons on the development set of \dn (-- means using the pre-trained language model for initialization).}
\end{table}

\noindent \textbf{Introducing more weakly-labeled data can lead to further performance gains.} Using the method mentioned in Section~\ref{sec:method:integration}, introducing additional weakly-labeled MRC instances generated based on verbal-nonverbal knowledge automatically extracted from scripts, we observe $+1.5\%$ in accuracy over the best-performing expert student (5 in Table~\ref{tab:performance}), which already outperforms the expert student obtained when we only use one-third of weakly-labeled data constructed based on ExamQA by $0.8\%$ in accuracy (Table~\ref{tab:more_data}). 
Furthermore, we show it is possible to use the same procedure to adapt self-teaching to incorporate \textbf{extra noisy human-labeled} multiple-choice MRC instances as there is a growing trend in constructing MRC benchmarks, especially in resource-rich languages such as English. We automatically translate instances from \dn's English counterparts RACE~\cite{lai2017race} and DREAM~\cite{sundream2018} that are also collected from language exams into Chinese (referred to as $\text{MRC}_\text{MT}$ in Table~\ref{tab:more_data}), and we apply self-teaching to additionally incorporate the data, leading to $+2.5\%$ in accuracy. We do not study how to further improve machine reading comprehension by just using extra \textbf{clean} human-annotated MRC data, which is not the main focus of this paper. These results suggest the flexibility and scalability of self-teaching, and increasing the amount of weakly-labeled data of different types may yield further improvements.

\begin{table}[h!]
\centering
\scriptsize
\begin{tabular}{lccc}
\toprule
\bf weakly-labeled data  & \bf size & \bf dev & \bf test \\
\midrule
--                            & --   &  73.9  \scriptsize(0.5)    & 73.4 \scriptsize(0.5) \\
subset of ExamQA              & 0.2M &  77.8 \scriptsize(0.2)      & 77.7 \scriptsize(0.1)\\
ExamQA                        & 0.6M &  78.2 \scriptsize(0.3)      & 78.5 \scriptsize(0.2)\\
ExamQA + SCRIPT              & 1.3M &   79.5 \scriptsize(0.2)      & 80.0 \scriptsize(0.2)\\
\midrule
\bf mixed-labeled data  &   &   & \\
ExamQA + $\text{MRC}_\text{MT}$ & 0.7M   &  \textbf{80.4} \scriptsize(0.1)       &   \textbf{81.0} \scriptsize(0.2) \\
\bottomrule
\end{tabular}
\caption{\label{tab:more_data} Accuracy comparison of expert students, which are obtained when different size of weakly-labeled MRC data is used during self-teaching, on the dev and test sets of the {\dn} dataset (size: number of instances).}
\end{table}

\subsection{The Usefulness of ExamQA for Extractive MRC Tasks}

We mostly follow the self-teaching paradigm introduced in Section~\ref{sec:method} and introduce how to apply self-teaching to extractive tasks by redefining hard and soft labels for probability distributions of being answer start and end tokens, changing the loss function for multi-skilled student and expert student from the original maximum likelihood to Kullback-Leibler divergence, etc., in Appendix~\ref{sec:appendix}. As there are major differences (\eg, type of questions/answers and required prior knowledge) between extractive and multiple-choice MRC tasks, we do not see positive results by adapting the resulting best-performing multiple-choice expert student to initialize an extractive model. 

As shown in Table~\ref{tab:cmrc}, similar to our observations on \dn, the expert student achieves the best performance, outperforming the baseline model we implemented based on the same pre-trained language model as previous work~\cite{cui-2020-revisiting} by $3.8\%$ in exact match and $2.0\%$ in F$1$. As each (question, document) corresponds to two probability distributions in a much larger dimension compared to that of soft labels for multiple-choice tasks, due to memory limitations, we only use the weakly-labeled extractive MRC data based on one-third of ExamQA instances (same as the subset of ExamQA in Table~\ref{tab:more_data}). It is likely that extractive MRC tasks will benefit from more weakly-labeled data.

\begin{table}[h!]
\centering
\scriptsize
\begin{tabular}{llll}
\toprule
\bf method                 &  \bf extra data      & \bf EM & \bf F1 \\
\midrule
\cite{cui-2020-revisiting} & N/A &  67.6  &  87.9  \\
\midrule
baseline                      & N/A &  70.3 \scriptsize(1.4)  &  89.2  \scriptsize(0.2) \\ %
multi-skilled teacher        & $\diamond$ &  71.8 \scriptsize(0.6)  &   89.8 \scriptsize(0.4)   \\
multi-skilled student        & $\diamond$ & 72.5 \scriptsize(0.6)      & 90.1    \scriptsize(0.5)  \\
expert student                                   & N/A &  \textbf{74.1} \scriptsize(0.7)   &   \textbf{91.2} \scriptsize(0.3) \\  %
\bottomrule
\end{tabular}
\caption{\label{tab:cmrc} EM and F1 (\%) on the publicly available development set of CMRC 2018 ($\diamond$: subset of ExamQA used for training multi-skilled teacher and multi-skilled student under self-teaching).}
\end{table}

We also explore the robustness of the resulting extractive MRC expert model in a relatively noisy setting: we consider an extractive MRC dataset DRCD~\cite{shao2018drcd} originally written in traditional Chinese wherein noise is caused by converting traditional characters into simplified ones, and we fine-tune the extractive expert model (Table~\ref{tab:cmrc}) on this converted dataset. We still see positive results in this noisy setting (Table~\ref{tab:drcd:dev} and~\ref{tab:drcd:test}).

\begin{table}[h!]
\centering
\scriptsize
\begin{tabular}{lll}
\toprule
\bf model initialization                       & \bf EM & \bf F1 \\
\midrule
\cite{cui-2020-revisiting}                     &  89.1  &  94.4  \\
\midrule
RoBERTa-wwm-ext-large                          &  90.5 \scriptsize(0.4)  &  95.3  \scriptsize(0.1) \\ 
baseline in Table~\ref{tab:cmrc}               &  90.4 \scriptsize(0.1)  &  95.2 \scriptsize(0.2) \\ 
expert student in Table~\ref{tab:cmrc}         &  \textbf{91.1} \scriptsize(0.2)  &  \textbf{95.7} \scriptsize(0.2)   \\ 
\bottomrule
\end{tabular}
\caption{\label{tab:drcd:dev} EM and F1 (\%) on the dev set of DRCD.}
\end{table}

\begin{table}[h!]
\centering
\scriptsize
\begin{tabular}{lll}
\toprule
\bf model initialization                       & \bf EM & \bf F1 \\
\midrule
\cite{cui-2020-revisiting}                     &  88.9  &  94.1  \\
\midrule
RoBERTa-wwm-ext-large                          &  90.2 \scriptsize(0.4)  &  94.9  \scriptsize(0.2) \\ 
baseline in Table~\ref{tab:cmrc}               &  90.5 \scriptsize(0.6)  &  94.9 \scriptsize(0.4) \\ 
expert student in Table~\ref{tab:cmrc}         &  \textbf{90.9} \scriptsize(0.1)  &  \textbf{95.2} \scriptsize(0.1)   \\ 
\bottomrule
\end{tabular}
\caption{\label{tab:drcd:test} EM and F1 (\%) on the test set of DRCD.}
\end{table}

\subsection{Comparing Self-Teaching and Multi-Teacher Paradigms}

\begin{table*}[h!]
\centering
\scriptsize
\begin{tabular}{lllccc}
\toprule
\bf paradigm    & \bf weakly-labeled data  & \bf data segmentation criteria & \bf \# of multi-skilled teachers  & \bf dev & \bf test \\
\midrule
self-teaching   & ExamQA & -- & 1 & \textbf{78.2} \scriptsize(0.3)    & \textbf{78.5} \scriptsize(0.2)    \\
multi-teacher   & ExamQA & random & 2  &  77.3 \scriptsize(0.5)   & 78.1 \scriptsize(0.2)     \\
multi-teacher   & ExamQA & random & 4  &  77.5 \scriptsize(0.5)   & 77.9 \scriptsize(0.2)     \\ 
\midrule
self-teaching   & SCRIPT & -- & 1  & 77.9 \scriptsize(0.4)    & 77.9 \scriptsize(0.4)    \\
multi-teacher   & SCRIPT & random & 4  &  77.7 \scriptsize(0.2)    & 77.5 \scriptsize(0.3)   \\
multi-teacher   & SCRIPT & knowledge type & 4  &  77.7 \scriptsize(0.4)    & 77.9 \scriptsize(0.3)   \\

\bottomrule
\end{tabular}
\caption{\label{tab:self-multi-teacher} Comparison of self-teaching and multi-teacher using different types of weakly-labeled data in accuracy (\%) on the dev and test sets of the {\dn} dataset.}
\end{table*}

For simplicity, we concentrate on multiple-choice tasks hereafter. Previous work shows that it is better to train multiple teacher models upon different types of weakly-labeled data with hard labels and then use these teachers to generate \textbf{soft} labels for both the weakly-labeled data and the small-scale MRC data, compared against training one model over the entire weakly-labeled data with \textbf{hard} labels and then fine-tuning it on the small-scale MRC dataset with \textbf{hard} labels~\cite{sun2020improving}. However, herein lies an unanswered question: \textbf{whether teacher models' data diversity or number matters to the resulting expert student's performance}.

As it is difficult to divide ExamQA into subsets by subjects, which can result in hundreds of teachers, we shuffle ExamQA and divide it into two subsets of similar size and follow the multi-teacher paradigm mentioned above. We compare it with the self-teaching paradigm and find that self-teaching provides larger accuracy gains compared against multi-teacher when knowledge-based data segmentation is tricky (Table~\ref{tab:self-multi-teacher}).

We also consider the case when it is easy to split data into subsets by the type of knowledge: we compare self-teaching with multi-teacher given the weakly-labeled data based on four types of verbal-nonverbal knowledge extracted from scripts. Results reveal that the two paradigms have similar performance, indicating that the impact of the number of teacher models may be \textbf{limited}. To study the impact of data diversity of teachers, we shuffle SCRIPT and divide it into four subsets of similar size to train four teacher models. Using the same multi-teacher paradigm, we experimentally demonstrate that there is a \textbf{weak} correlation between the data diversity of teachers and the final performance of the expert student.

\subsection{The Impact of Cleanness and Source of Context in Weakly-Labeled Data}

\begin{table}[h!]
\centering
\scriptsize
\begin{tabular}{lccc}
\toprule
\bf source of context  & \bf denoise & \bf dev & \bf test \\
\midrule
search engine             & $\times$ &   \textbf{78.2} \scriptsize(0.3)     & \textbf{78.5} \scriptsize(0.2)\\
search engine             & $\checkmark$ &  77.0 \scriptsize(0.3)       & 77.5 \scriptsize(0.3)\\
Wikipedia                 & $\times$ &  77.1  \scriptsize(0.3)      &  77.4 \scriptsize(0.2)\\
\bottomrule
\end{tabular}
\caption{\label{tab:context} Accuracy comparison of expert students on the dev and test sets of the {\dn} dataset, which are obtained when different types of sources are used to form context of weakly-labeled data.}
\end{table}

As mentioned previously, context returned by a web search engine tends to be noisy. For example, given a question as the search query, the question and its wrong answer options are included in context. We conduct a preliminary experiment to evaluate the impact of context cleanliness by removing wrong answer options from the context of each weakly-labeled MRC instance. However, context cleaning \textbf{hurts} accuracy by $1.2\%$ on the development set of \dn. It is possible that noisy context helps improve the generalization ability of both teacher and student models, similar to the role of noise (\eg, dropout and data augmentation) that is intentionally injected to the student models during self-training in previous studies (\eg,~\cite{he2019revisiting,xie2020self}).

Besides using snippets retrieved from a search engine to form context, we use the default search engine in Wikipedia to collect relevant snippets from Wikipedia for each question, leading to decreased accuracy ($-1.1\%$ on \dn), perhaps due to answering questions in ExamQA requires fine-grained subject-specific knowledge that is seldom covered in Wikipedia. In the future, we are interested in using different types of texts (\eg, news reports and dialogues) as context of structured knowledge such as question-answering instances to study the impact of relevance, completeness, and style on performance on downstream tasks.

%% file: 5_related_work.tex
\section{Related Work}

\subsection{From Question Answering to Machine Reading Comprehension}

Here we do not compare with one direction in transfer learning in MRC when the source and target tasks are all clean MRC tasks~\cite{chung-etal-2018-supervised,wang-2018-yuanfudao,shakeri-etal-2020-end,nishida-etal-2020-unsupervised}, as it is expensive and time-consuming to construct high-quality large-scale MRC datasets considering factors such as ensuring the high relevance between questions and documents and the degree of difficulty of questions.

This work is related to data augmentation in semi-supervised MRC studies, which partially or fully rely on the document-question-answer triples~\cite{yang-etal-2017-semi,yuan2017machine,yu2018qanet,zhang-bansal-2019-addressing,zhu-2019-learning,dong2019unified,sun2018improving,alberti-etal-2019-synthetic,asai-hajishirzi-2020-logic,rennie-2020-unsupervised} of target MRC tasks or at least similar domain corpora~\cite{dhingra-etal-2018-simple}. We mainly focus on studying leveraging multi-domain question-answering data to improve different types of general-domain MRC tasks; though, at first glance, subject-area knowledge is seldom required for these MRC tasks.

To the best of our knowledge, ExamQA is the largest multi-subject QA dataset collected from exams to date. We offer ExamQA mainly for the purpose of using large-scale subject-area QA data to improve other tasks such as MRC, rather than focusing on improving single-subject or multi-subject question answering.

\subsection{Teacher-Student Paradigms}

Teacher-student paradigms are widely used for knowledge distillation~\cite{ba2014deep,li2014learning,hinton2015distilling}. We aim to let the student model outperform its teacher model to improve existing competitive supervised methods and simply use the same architecture for all teacher and student models.

Our work is related to self-training~\cite{yarowsky-1995-unsupervised,riloff1996automatically}, as we also leverage unstructured texts for context generation. The main difference is that we generate weakly-labeled data based on existing large-scale QA data covering a wide range of domains, instead of the same domain~\cite{he2019revisiting,xie2020unsupervised,zhao-etal-2020-robust,chen2020improved} or at least approximately in-domain~\cite{du2020self} as the target task. Different from previous studies that iteratively use new teacher models to generate new pseudo data from unlabeled data (\eg, ~\cite{wang2020combining}), we use the new teacher model to generate new soft labels for fixed weakly-labeled and target data. Furthermore, we use a search engine to retrieve noisy and incomplete snippets instead of full sentences or even documents, which are seldom studied and used as context for structured knowledge via distant supervision for downstream natural language understanding tasks~\cite{ye2019align}.

Compared with previous multi-teacher student paradigms~\cite{you2019teach,wang2019go,yang2020model}, we conduct iterative training and leverage large-scale weakly-labeled data to train models to be teachers, instead of using human-labeled clean data of similar tasks.

%% file: 6_conclusion.tex
\section{Conclusions}

It is still unclear how to improve MRC using large-scale subject-area QA data. In this paper, we collect a large-scale multi-subject multiple-choice QA dataset ExamQA. We use incomplete and noisy snippets returned by a web search engine as relevant context of each QA instance to convert it into a weakly-labeled MRC instance. Furthermore, we propose a self-teaching paradigm to better use these weakly-labeled MRC instances to improve an MRC task of interest. Experimental results show that we can obtain $+5.1\%$ in accuracy on a multiple-choice MRC dataset \dn~and $+3.8\%$ in exact match on an extractive MRC dataset CMRC 2018, demonstrating the effectiveness of our framework and the usefulness of large-scale subject-area QA data for MRC. 

%% file: appendix.tex
\section{Appendix}
\label{sec:appendix}

\subsection{Training a Multi-Skilled Teacher}
\label{appendx:method:multi-teacher}

Let $V$ denote a set of human-labeled instances and $W$ denote a set of weakly-labeled instances. Each instance contains a document $d$, a question $q$, and an answer span $a$ in $d$. Let $a_{\text{start}}$ and $a_{\text{end}}$ denote, respectively, the start offset and end offset of $a$, which appears in $d$. For each instance $t$ = ($d$, $q$, $a$), let $l_t$ denote the length of the concatenated ($q$, $d$) taken as the input to an MRC model.

We train a multi-skilled teacher model, denoted by $\mathcal{T}$, which learns to predict the probability of each token in the input to be the start or end token of the correct answer. 
Let $p_{\text{start}, \theta}(k\,|\,t)$ and $p_{\text{end}, \theta}(k\,|\,t)$ denote the probabilities that the $k$-th token in ($q$, $d$) to be the start and end token respectively, estimated by a model with parameters $\theta$. We optimize $\mathcal{T}$ by minimizing $\sum_{t \in V\cup W} L_1(t, \theta_{\mathcal{T}})$, where $L_1$ is defined as 

{\small
\begin{equation*}
L_{1}(t, \theta) = - \log p_{\text{start}, \theta}(a_{\text{start}}\,|\,t)  - \log p_{\text{end}, \theta}(a_{\text{end}}\,|\,t). 
\end{equation*}
}%

\subsection{Training a Multi-Skilled Student}
\label{appendix:method:multi-student}

We then train a multi-skilled student model $\mathcal{S}$ using the same data as the multi-skilled teacher model $\mathcal{T}$ while replacing the hard labels with the soft labels predicted by $\mathcal{T}$. We define ${\bm h}^{(t)}_{\text{start}}$ and ${\bm h}^{(t)}_{
\text{end}}$ to be one-hot hard-label vectors such that ${\bm h}^{(t)}_{\text{start},i} = 1$ and ${\bm h}^{(t)}_{\text{end},j}=1$ if the $i$-th and $j$-th tokens in ($q$, $d$) are the start and end token of the correct answer respectively. We define soft-label vectors ${\bm s_\text{start}}^{(t)}$ and ${\bm s_\text{end}}^{(t)}$ for $t \in V\cup W$ such that
{\small
\begin{equation*}
    {\bm s}^{(t)}_{\text{start},k}  =
     \lambda~{\bm h}^{(t)}_{\text{start}, k} + (1 - \lambda) p_{\text{start}, \theta_{\mathcal T}}(k\,|\,t)  \\
\end{equation*}
}%
and 
{\small
\begin{equation*}
    {\bm s}^{(t)}_{\text{end},k}  =
     \lambda~{\bm h}^{(t)}_{\text{end},k} + (1 - \lambda) p_{\text{end}, \theta_{\mathcal T}}(k\,|\,t)  \\,
\end{equation*}
}%
where $\lambda\in [0, 1]$ is a weight parameter, and $k = 1,\dots, l_t$. We optimize multi-skilled student $\mathcal{S}$ by minimizing $\sum_{t \in V\cup W}L_2(t, \theta_{\mathcal S})$, where $L_2$ is defined as
{\small
\begin{align*}
     L_{\text{start}, 2}(t, \theta) & = -\sum_{1\le k\le l_t} {\bm{s}}^{(t)}_{\text{start}, k} ~ \log p_{\text{start}, \theta}(k\,|\,t) \\
     L_{\text{end}, 2}(t, \theta) & = -\sum_{1\le k\le l_t} {\bm{s}}^{(t)}_{\text{end},k} ~ \log p_{\text{end}, \theta}(k\,|\,t)  \\
     L_{2}(t, \theta) & = \frac{1}{2} (L_{\text{start}, 2}(t, \theta) + L_{\text{end}, 2}(t, \theta) ) .
\end{align*}
}%

\subsection{Training an Expert Student}
\label{appendix:method:expert-student}

We now introduce the formulation of training expert student $\mathcal{E}$. For instance $t\in V$, we define new soft-label vectors ${\bm{\tilde{s}_\text{start}}}^{(t)}$ and ${\bm{\tilde{s}_\text{end}}}^{(t)}$ such that 
{\small
\begin{equation*}
    { \bm{\tilde s}}^{(t)}_{\text{start},k} =
     \lambda~{\bm h}^{(t)}_{\text{start},k} + (1 - \lambda) p_{\text{start}, \theta_{\mathcal S}}(k\,|\,t)  \\ 
\end{equation*}}%
and
{\small
\begin{equation*}
    { \bm{\tilde s}}^{(t)}_{\text{end},k} =
     \lambda~{\bm h}^{(t)}_{\text{end},k} + (1 - \lambda) p_{\text{end}, \theta_{\mathcal S}}(k\,|\,t)  \\,
\end{equation*}}%
where $\lambda\in [0, 1]$ is a weight parameter, and $k = 1,\dots, l_t$. We optimize $\mathcal{E}$ by minimizing $\sum_{t\in V} L_3(t, \theta_{\mathcal E})$, where $L_3$ is defined as
{\small
\begin{align*}
     L_{\text{start}, 3}(t, \theta) & = -\sum_{1\le k\le l_t} {\bm{\tilde{s}}}^{(t)}_{\text{start},k} ~ \log p_{\text{start}, \theta}(k\,|\,t) \\
     L_{\text{end}, 3}(t, \theta) & = -\sum_{1\le k\le l_t} {\bm{\tilde{s}}}^{(t)}_{\text{end},k} ~ \log p_{\text{end}, \theta}(k\,|\,t)  \\
     L_{3}(t, \theta) & = \frac{1}{2} (L_{\text{start}}(t, \theta) + L_{\text{end}}(t, \theta) ) .
\end{align*}
}%

\subsection{Settings}
\label{sec:appendix:settings}

\begin{table}[h]
\centering
\scriptsize
\begin{tabular}{lll}
\toprule
                      & \bf mst/mss & \bf es/baseline \\
\midrule
training data         &  ExamQA + \dn                 & \dn~                               \\
initial learning rate &   2e-5                            & 2e-5                     \\
batch size            &   24                               &   24                    \\
\# of training epochs       &    1                       &      8                           \\
max sequence length         &  512                         & 512                               \\
training labels                    & hard/soft          & soft/hard   \\
\bottomrule
\end{tabular}
\caption{\label{tab:hyper} Hyper-parameter settings for training multiple-choice machine reading comprehension models (mst: multi-skilled teacher; mss: multi-skilled student; es: expert student).}
\end{table}

\begin{table}[h]
\centering
\scriptsize
\begin{tabular}{p{2.4cm}p{1.9cm}p{2.1cm}}
\toprule
                      & \bf mst/mss  & \bf es/baseline \\
\midrule
training data                           & $\diamond$  + CMRC 2018   & CMRC 2018 / DRCD                 \\
initial learning rate                   &  3e-5          & 3e-5      \\
batch size                                &  32     &  32   \\
\# of training epochs                      &  1     &     2         \\
max sequence length                         & 512   & 512           \\
training labels                                  & hard/soft          & soft/hard  \\
\bottomrule
\end{tabular}
\caption{\label{tab:hyper2} Hyper-parameter settings for training extractive machine reading comprehension models ($\diamond$: subset of ExamQA; mst: multi-skilled teacher; mss: multi-skilled student; es: expert student).}
\end{table}